\documentclass[11pt]{article}
\usepackage[preprint]{acl}

\usepackage{times}
\usepackage{latexsym}
\usepackage[T1]{fontenc}
\usepackage[utf8]{inputenc}
\usepackage{microtype}
\usepackage{inconsolata}
\usepackage{graphicx}
\usepackage{booktabs}
\usepackage{amsmath}
\usepackage{amssymb}
\usepackage{xcolor}
\usepackage{array}

\title{Crayotter: Traceable Multi-Agent Workflows for Long-Form Video Editing}

\author{
  Lecheng Yan$^{1*\dagger}$,
  Yichong Zhang$^{2*\dagger}$,
  Xiantao Xu$^{3*}$,
  Jianze Lin$^{4*}$,
  Ben Pan$^{5*}$, \\
  \bf{Xiaoyu Zheng$^{6*}$, 
  Jiawei Qian$^{6*}$},
  Anqi Wu$^{6*}$,
  Jiahui Geng$^{7}$, Ruizhe Li$^{8}$, \\ 
  \bf{Fengyu Cai$^{9}$, Jingcheng Niu$^{9}$, Raymond Li$^{10}$,}
  Wenxi Li$^{6}$,
  Chenyang Lyu$^{11}$ \\
  $^{1}$University of Science and Technology of China 
  $^{2}$Jilin University \\
  $^{3}$Dmall Inc. 
  $^{4}$Beijing Normal University 
  $^{5}$Tianjin University \\
  $^{6}$East China Normal University 
  $^{7}$Linköping University \\
  $^{8}$University of Birmingham 
  $^{9}$Technische Universität Darmstadt \\
  $^{10}$University of British Columbia 
  $^{11}$Alibaba Group \\
  \texttt{lyuchenyang.dcu@gmail.com}
}

\begin{document}
\maketitle

\begingroup
\renewcommand{\thefootnote}{\fnsymbol{footnote}}
% \footnotetext[1]{Core Contributor.}
% \footnotetext[2]{Project Lead.}
\footnotetext[1]{%
  Core Contributor.\quad
  \textsuperscript{\textdagger}\,Project Lead.%
}
\footnotetext[3]{Our code is publicly available at \url{https://github.com/idwts/Crayotter} and our live demo at \url{https://idwts.github.io/Crayotter/\#demo}.}
\endgroup

\begin{abstract}
Long-form video editing over heterogeneous footage requires agents to coordinate source selection, multimodal analysis, timeline construction, narration and subtitle alignment, rendering, and revision while exposing intermediate state for inspection and repair. We present \textbf{Crayotter}, an open-source multimodal multi-agent demo system for prompt-driven long-form video editing. Crayotter organizes production around coverage-aware material preparation, artifact-grounded editing research, and tool-grounded timeline execution. Across these stages, retrieval reports, video analyses, editing blueprints, scheduler events, tool calls, intermediate renders, and final exports are treated as first-class artifacts rather than hidden transient state. The workbench supports local assets, agent-assisted retrieval, progress monitoring, artifact preview, failure diagnosis, interrupted-job resumption, and resource-aware asynchronous execution for long-running workflows. In a 23-theme evaluation, Crayotter achieves the highest human overall score (3.40/5) among the compared systems, with its largest margins in theme alignment, narrative coherence, and editing smoothness. These results show that long-horizon video editing agents can be made traceable, inspectable, and practically controllable through observable production artifacts.
% Code, traces, and examples are publicly available at \url{https://github.com/idwts/Crayotter}.
\end{abstract}

\section{Introduction}

Long-form video editing transforms a high-level brief and heterogeneous footage into a coherent audiovisual narrative. Unlike short-form generation, it requires joint decisions over material selection, clip understanding, narrative structure, timeline operations, and audio--text alignment. LLM agents enable interleaved reasoning and tool use \cite{react23} and role-structured collaboration \cite{autogen23,metagpt23}; visual-production systems extend these capabilities to planning and specialized filmmaking roles \cite{videoDirectorGPT24,filmAgent25}. However, they primarily synthesize shots, whereas editing real or retrieved footage requires source-grounded decisions and executable timecodes.

Editing agents increasingly support hierarchical media compilation \cite{li2026direct} and long-horizon, music-synchronized execution \cite{Zhao2026CutClaw}. Practical systems must also expose why footage was selected, how operations changed the timeline, and where failures occurred. Without such evidence, users cannot inspect intermediate decisions, revise local segments, or resume interrupted runs. We therefore formulate prompt-driven long-form editing as a \emph{traceable} agent workflow whose production state is observable and reusable.

\begin{figure*}[t!]
  \centering
  \includegraphics[width=\textwidth]{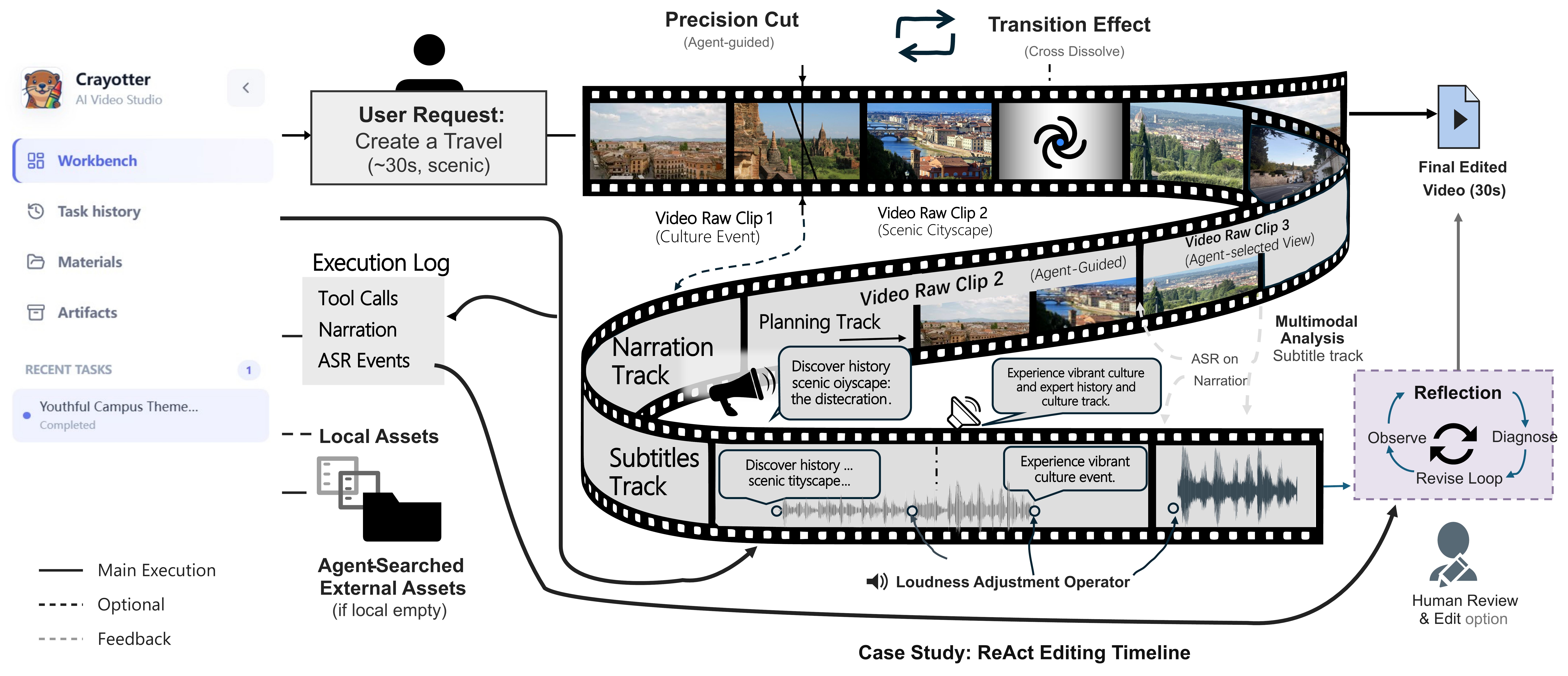}
\caption{Crayotter workbench and an editing trajectory. From left to right: task and asset specification; tool-grounded timeline construction for a travel-video request; and artifact-based review and revision before export.}
\label{fig:analysis}
\end{figure*}

We present \textbf{Crayotter}, an open-source multimodal multi-agent demo system. Unlike pipelines that assume fixed inputs or synthetic assets, Crayotter automatically searches multiple video sources when local material is insufficient. It constructs a coverage-verified material pool, derives a time-grounded blueprint, and executes it through registered timeline tools. Retrieval reports, analyses, plans, tool events, and intermediate renders remain first-class artifacts. Building on artifact-centered research and presentation systems \cite{researStudio25,deepPresenter26}, the workbench exposes them for inspection, checkpointed resumption, and localized re-execution, as illustrated in Figure~\ref{fig:analysis}.

Across 23 editing themes, Crayotter obtains the highest aggregate scores under both human and GPT-5.4 evaluation, including a human score of 3.40/5. Its resource-aware asynchronous runtime further improves end-to-end throughput by approximately $1.6\times$ over the previous scheduling path.

Our contributions are three-fold:
\begin{enumerate}
\item We formulate prompt-driven long-form video editing as a traceable workflow whose source evidence, plans, timeline states, renders, and logs support inspection and localized repair.
\item We develop a three-phase multi-agent system that integrates coverage-aware retrieval, time-grounded planning, and tool-grounded execution within an observable and resumable workbench.
\item We release the code, live demo, and execution traces; evaluate the system across 23 themes with human and AI judges; and introduce a resource-aware runtime for long-running production.
\end{enumerate}

\begin{figure*}[t]
  \centering
  \makebox[\columnwidth][c]{\includegraphics[width=.98\textwidth]{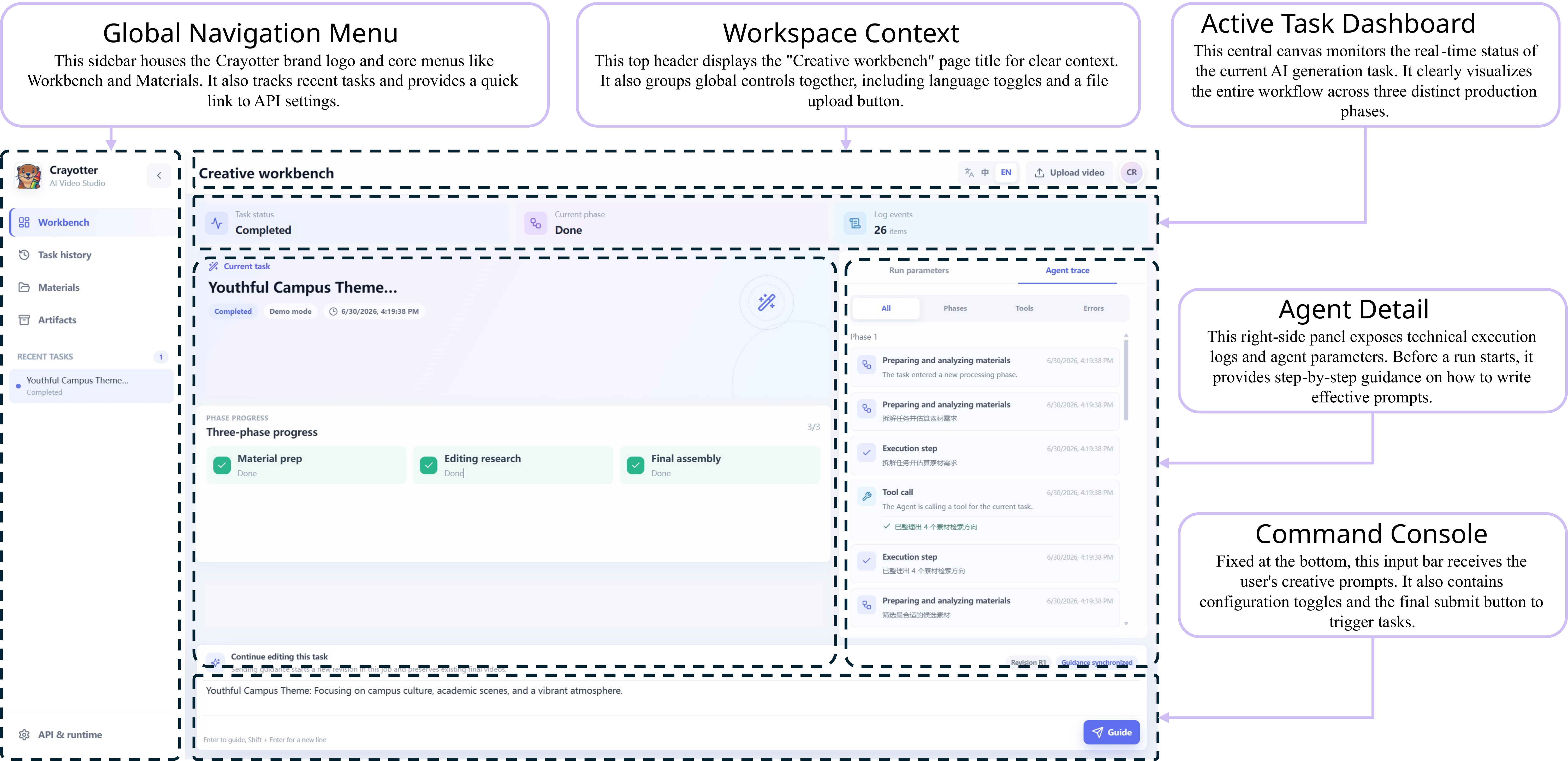}}
\caption{Crayotter workbench interface. The workspace exposes global navigation, task context and status, agent details, command logs, and intermediate artifacts for monitoring and inspecting long-running editing workflows.}
  \label{fig:ui}
\end{figure*}

\section{Related Work}

\textbf{Agentic visual production.} VideoDirectorGPT decomposes multi-scene generation through LLM-based planning \cite{videoDirectorGPT24}. DreamFactory and MovieAgent extend role-based coordination to long-form narrative production \cite{dreamFactory24,movieAgent25}. MM-StoryAgent integrates text, image, and audio for narrated storybooks \cite{mmStoryAgent25}, whereas FilmAgent simulates specialized filmmaking roles in virtual 3D environments \cite{filmAgent25}. AniMaker further introduces MCTS-based clip exploration for animated storytelling \cite{animaker25}. These systems establish effective agent specialization for content synthesis; Crayotter instead coordinates post-production decisions over heterogeneous source footage and preserves the corresponding executable tool trajectory.

\textbf{Long-form video generation and editing.} FIFO-Diffusion extends video duration through an autoregressive diffusion process \cite{fifo24}, while Long Context Tuning adapts video models to longer temporal contexts \cite{lct25}. MinT targets explicit multi-event temporal control \cite{mint25}, and test-time training supports minute-scale generation \cite{oneMinute25}. Story-visualization research studies multi-subject consistency and iterative refinement \cite{dreamStory24,storyIter24}; complementary benchmarks evaluate story-level consistency \cite{viStoryBench26} and generated-video quality \cite{evalCrafter24}. Closer to our setting, DIRECT performs hierarchical, intent-guided mashup creation \cite{li2026direct}, CineAgents studies instruction-driven cinematic compilation \cite{zhang2026benchmark}, and CutClaw targets music-synchronized hours-long editing \cite{Zhao2026CutClaw}. FireRed-OpenStoryline combines conversational media search, planning, tool orchestration, and human control,\footnote{\url{https://github.com/FireRedTeam/FireRed-OpenStoryline}} while NarratoAI automates script, narration, and video assembly for film commentary.\footnote{\url{https://github.com/linyqh/NarratoAI}} Crayotter complements these systems with coverage evidence, time-grounded blueprints, and localized failure traces throughout execution.

\textbf{Intervenable agent workflows.} ResearStudio externalizes plans, file changes, and tool activity to support real-time user intervention \cite{researStudio25}. DeepPresenter grounds iterative refinement in rendered presentation artifacts \cite{deepPresenter26}. Crayotter transfers this artifact-centered principle to temporal video editing, where users must inspect clip evidence, time-grounded plans, timeline operations, and render diagnostics. Its contribution therefore lies not only in the final video, but also in an executable production trajectory that can be audited, resumed, and selectively repaired.

\section{System Design: Crayotter Demo Architecture}

The complete system architecture and artifact flow are illustrated in Appendix~\ref{app:architecture} (Figure~\ref{fig:overview}).   Crayotter maps a user request $q$ and optional local assets $\mathcal{A}_0$ to an exported video $y$ through three artifact-producing phases. Phase~1 constructs a coverage-verified material pool $\mathcal{P}=(\mathcal{S},\Gamma,\mathcal{M})$, where $\mathcal{S}$ contains selected videos, $\Gamma$ records coverage evidence, and $\mathcal{M}$ stores multimodal analyses. Phase~2 converts $\mathcal{P}$ into a time-grounded editing blueprint $\mathcal{B}$, and Phase~3 realizes that blueprint as an observable execution history $\mathcal{H}$ and final render.

\begingroup
\[
(q,\mathcal{A}_0)
\xrightarrow{\text{Phase 1}} \mathcal{P}
\xrightarrow{\text{Phase 2}} \mathcal{B}
\xrightarrow{\text{Phase 3}} (\mathcal{H},y).
\]
\endgroup

The decomposition exposes the information required for inspection and repair rather than retaining it in a latent agent state. Specifically, $\mathcal{B}=\{b_m\}_{m=1}^{M_z}$ contains shot-level decisions, while $\mathcal{H}=\{(s_k,a_k,s_{k+1},\zeta_k)\}_{k=1}^{K}$ records each tool action and its artifact-level diagnostic $\zeta_k$.

Figure~\ref{fig:ui} shows the corresponding workbench. Users may upload source clips or invoke agent-assisted retrieval, submit an editing brief, monitor phase-level progress, inspect intermediate artifacts, and access the final export within a single workspace. Interrupted jobs retain validated artifacts and scheduler checkpoints, enabling execution to resume without reconstructing the complete trajectory.

\begin{figure*}[t]
  \centering
  \makebox[\columnwidth][c]{\includegraphics[width=.98\textwidth]{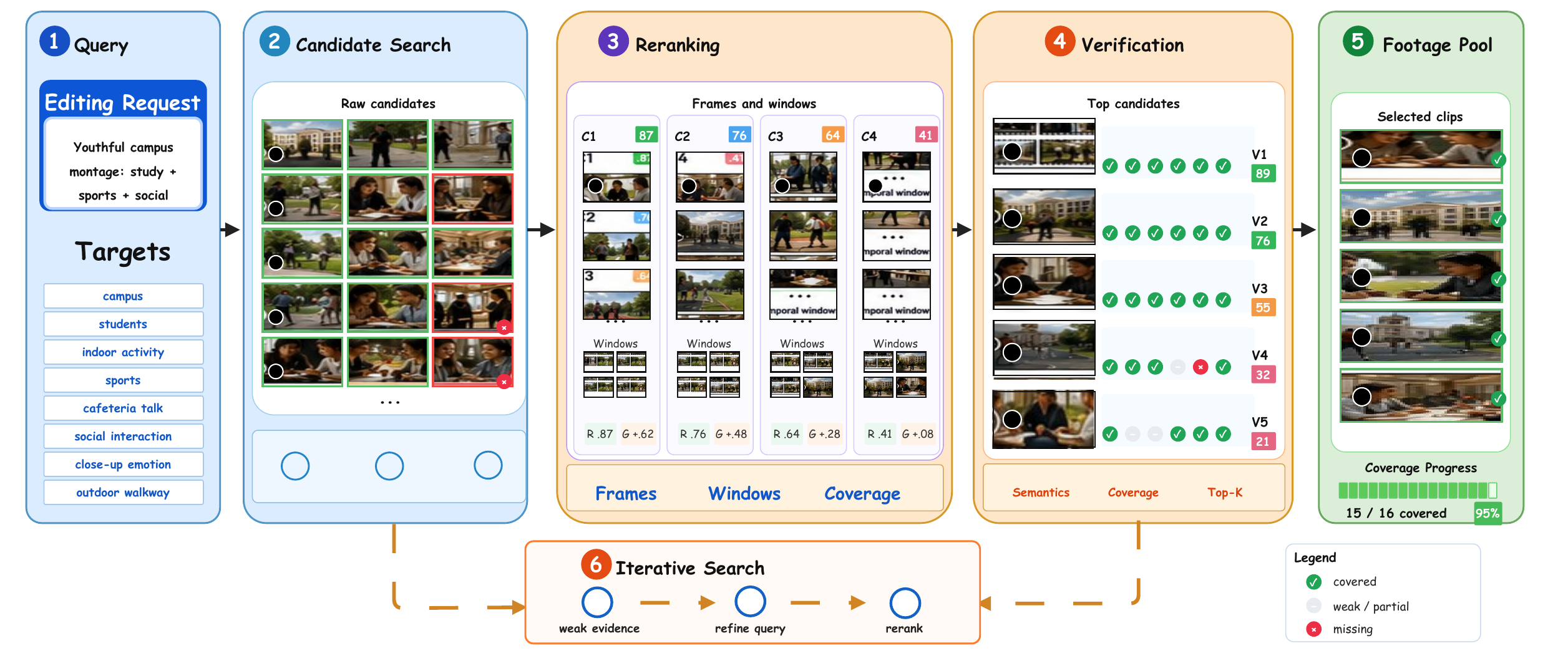}}
\caption{Coverage-aware multimodal footage retrieval. Crayotter expands an editing request into production tags, retrieves a high-recall candidate pool, reranks candidates using frame- and temporal-window evidence, verifies top-ranked videos with full-video analysis, and issues follow-up queries for uncovered tags until coverage is sufficient or the retrieval budget is exhausted.}
  \label{fig:retrieval}
\end{figure*}

\subsection{Phase 1: Coverage-Aware Multimodal Footage Retrieval}

Effective timeline planning requires source material that covers the requested scenes, actions, narrative functions, shot types, and styles. Phase~1 therefore expands $q$ into weighted production tags $T=\{(t_i,w_i)\}$ and retrieves a high-recall candidate set. For each candidate $v$, the multimodal analyzer estimates tag support $g(v,t_i)$ from sampled frames and temporal windows. Let $G_{\mathcal{S}}(t_i)=\max_{v\in\mathcal{S}}g(v,t_i)$ denote the evidence already supplied by the selected pool. Candidate selection combines request relevance with the marginal evidence it contributes:

\begingroup
\[
\begin{aligned}
D_i(v,\mathcal{S})
&=\max\!\left(0,g(v,t_i)-G_{\mathcal{S}}(t_i)\right),\\
\operatorname{score}(v\mid\mathcal{S},T)
&=\lambda s_{\mathrm{rel}}(q,v)\\
&\quad +(1-\lambda)\sum_i w_iD_i(v,\mathcal{S}),\\
\operatorname{Cov}(\mathcal{S},T)
&=\frac{\sum_i w_i\mathbb{I}[G_{\mathcal{S}}(t_i)\ge\eta_i]}
{\sum_i w_i}.
\end{aligned}
\]
\endgroup

As illustrated in Figure~\ref{fig:retrieval}, top-ranked videos undergo full-video verification before entering $\mathcal{S}$. Tags below their support thresholds form a coverage gap $\Delta_r$ that conditions the next retrieval query. The loop terminates when coverage is sufficient or the retrieval budget is exhausted, yielding selected assets $\mathcal{S}$, tag-level evidence $\Gamma$, and per-video analyses $\mathcal{M}$.

\subsection{Phase 2: Deep Editing Research}

Phase~2 performs planning without invoking editing tools. Its principal task is temporal localization: each narrative beat must be associated with both a source video and an executable in/out interval. Crayotter overlays human-readable temporal coordinates on sampled evidence, producing $\tilde{v}_j=R_{\mathrm{time}}(v_j)$. This interface binds semantic observations to absolute source timestamps without modifying the underlying multimodal model.

For narrative beats $\mathcal{Z}=\{z_m\}_{m=1}^{M_z}$ derived from $q$ and $\Gamma$, the research agent constructs a structured blueprint:

\begingroup
\[
\begin{aligned}
b_m&=(v_{\hat{j}_m},\hat{\tau}_m,c_m,\delta_m,n_m),\\
\mathcal{B}&=\{b_m\}_{m=1}^{M_z}.
\end{aligned}
\]
\endgroup

Here $\hat{\tau}_m$ is the selected interval, $c_m$ specifies its cinematic role, $\delta_m$ encodes transition and pacing intent, and $n_m$ specifies narration or subtitle intent. The resulting plan is temporally addressable rather than free-form: users and downstream tools can inspect, revise, or execute each beat independently.

\subsection{Phase 3: Tool-Grounded Timeline Execution}

Phase~3 translates $\mathcal{B}$ into concrete timeline operations. State $s_k$ comprises the current timeline, media assets, narration and subtitle tracks, rendered previews, and tool logs; an action $a_k$ applies a registered editing operator such as trimming, transition insertion, narration alignment, loudness normalization, or export. Execution and diagnosis are coupled as follows:

\begingroup
\[
\begin{aligned}
s_{k+1}&=E(s_k,a_k;\mathcal{B}),\\
\zeta_k&=D_{\mathrm{tool}}(s_k,a_k,s_{k+1},\mathcal{B},q).
\end{aligned}
\]
\endgroup

The diagnostic $\zeta_k$ summarizes observable properties including tool validity, timestamp accuracy, coverage preservation, narration alignment, transition continuity, and render quality. When a check fails, the agent grounds reflection in the affected artifact and re-executes the corresponding segment, rather than restarting the complete edit.

\subsection{Agent Roles and Artifact Contracts}

The implementation separates planner, researcher, and executor responsibilities. The planner derives phase goals and coverage requirements; the researcher synthesizes multimodal evidence into $\mathcal{B}$; and the executor converts blueprint entries into validated tool calls. Their interface is a stable artifact contract spanning input evidence, planning outputs, and execution products such as timelines, intermediate renders, and logs. Before narration, the merged video undergoes composite re-analysis so that scripts are conditioned on rendered content rather than source analyses alone. This contract supports replay, phase-level diagnosis, and selective re-execution.

\subsection{Modular Tool Taxonomy}

Crayotter exposes 23 registered tools for material preparation and timeline execution. Table~\ref{tab:tools} presents the complete functional taxonomy.

\begin{table}[t]
\caption{Tool taxonomy with functional grouping.}
\label{tab:tools}
\centering
\small
\resizebox{\columnwidth}{!}{
\begin{tabular}{ll}
\toprule
Category & Tools \\
\midrule
Search & \texttt{search\_video} \\
Ranking & \texttt{rank\_video\_candidates} \\
Download & \texttt{download\_video} \\
Analysis & \texttt{analyze\_video}, \texttt{recall\_semantic\_segments} \\
Cutting & \texttt{cut\_video}, \texttt{batch\_cut\_video} \\
Merging & \texttt{merge\_videos} \\
Inspection & \texttt{inspect\_video\_duration} \\
Transitions & \texttt{add\_transition}, \texttt{plan\_transition\_timeline}, \\
 & \texttt{list\_transition\_presets} \\
Continuity & \texttt{score\_cut\_continuity}, \\
 & \texttt{recommend\_transition\_for\_cut} \\
Timeline & \texttt{build\_edit\_timeline\_from\_segments}, \\
 & \texttt{align\_narration\_to\_timeline}, \\
 & \texttt{validate\_timeline\_constraints} \\
Narration & \texttt{add\_narration\_segments}, \\
 & \texttt{validate\_narration\_timeline} \\
Audio Post & \texttt{duck\_background\_audio}, \texttt{normalize\_loudness} \\
Subtitles & \texttt{add\_subtitles} \\
Export & \texttt{export\_video} \\
\bottomrule
\end{tabular}
}
\end{table}

The transition subsystem provides configurable basic, motion, and cinematic effects, while continuity tools estimate visual discrepancies around cut points and recommend an appropriate transition. Phase routing supports full, local-first, and direct-execution modes. These components share the same artifact contract, allowing their outputs and validation signals to participate in the reflection loop without altering the three-phase abstraction.

\subsection{Asynchronous Scheduling and Demo Runtime}

Long-horizon editing contains substantial parallelism but also shared-write constraints. Crayotter represents each phase as a resource-aware task DAG whose nodes declare dependencies, required resources, output artifacts, conflicts, and retry policies. The scheduler validates the graph, executes independent search, download, analysis, research, cutting, and narration tasks with bounded concurrency, and serializes operations that modify shared timelines or final outputs.

Resource limits are configurable from the workbench to accommodate different hardware profiles. Relative to the previous scheduling path, the asynchronous runtime improves end-to-end throughput by approximately 1.6$\times$ by overlapping model and network latency with independent media processing while preserving dependency and write-conflict constraints.

The scheduler checkpoints task state and validates produced artifacts before reuse. An interrupted job can therefore resume from the latest valid checkpoint; changed or missing dependencies invalidate only the affected tasks. The workbench streams phase progress, task events, artifacts, and failures, making runtime state available for inspection and localized recovery.

\section{Evaluation}

\subsection{Evaluation Setup}

We compare Crayotter with three baselines: CapCut-Mate, CutClaw, and ChatCut, a proprietary prompt-based browser editor.\footnote{\url{https://chatcut.io}} ChatCut was run with proprietary GPT-5.5, while the open-source Crayotter used Qwen3.5-122B-A10B \cite{qwen35}. The benchmark includes prompt metadata and final videos for all methods and, for Crayotter, configuration files and tool traces for process-level inspection. The themes span five scenario families: pet (3), campus (5), travel (5), scenery (5), and food (5). Each output is rated on a five-point scale for theme alignment, content richness, narrative coherence, editing smoothness, and visual quality; the overall score is the mean across dimensions.

\subsection{Scoring Protocol}

We report both human judgments and GPT-5.4 judge scores. For GPT-5.4 scoring, we use the same multidimensional rubric across all 23 themes and all methods. For human scoring, three annotators independently score the same outputs; we first average their scores for each theme--method pair, then average across the 23 themes. This case-level aggregation avoids giving extra weight to any annotator. Inter-annotator reliability is measured using a two-way random-effects, absolute-agreement intraclass correlation coefficient for the mean of three annotators, ICC(A,3) \citep{koo2016guideline}. Across the 92 case--method outputs, the overall-score ICC is 0.86 (95\% cluster-bootstrap CI: 0.78--0.91), while dimension-level ICCs range from 0.77 to 0.83. Table~\ref{tab:eval_summary} reports aggregate results; Appendix~\ref{app:case_scores} provides case-level human and GPT-5.4 scores for Crayotter, CapCut-Mate, CutClaw, and ChatCut.

\begin{table*}[!t]
\caption{Aggregate evaluation on 23 themes. Dimension columns report mean scores on a 1--5 scale; Overall reports mean $\pm$ standard deviation across themes. GPT-5.4 uses the same rubric as human evaluation.}
\label{tab:eval_summary}
\centering
\resizebox{0.98\textwidth}{!}{
\begin{tabular}{llcccccc}
\toprule
\textbf{Judge} & \textbf{Method} & \textbf{Theme} & \textbf{Richness} & \textbf{Narrative} & \textbf{Smoothness} & \textbf{Visual} & \textbf{Overall} \\
\midrule
Human & \textbf{Crayotter} & \textbf{3.59} & \textbf{3.36} & \textbf{3.22} & \textbf{3.29} & \textbf{3.54} & $\mathbf{3.40}\pm0.59$ \\
Human & ChatCut & 3.42 & 3.22 & 3.03 & 3.14 & 3.35 & $3.23\pm0.75$ \\
Human & CapCut-Mate & 2.59 & 2.71 & 2.01 & 2.13 & 2.74 & $2.44\pm0.72$ \\
Human & CutClaw & 1.59 & 1.64 & 1.72 & 1.86 & 1.67 & $1.70\pm0.62$ \\
\midrule
GPT-5.4 & \textbf{Crayotter} & \textbf{2.78} & 2.61 & \textbf{2.13} & \textbf{2.17} & 2.26 & $\mathbf{2.39}\pm0.62$ \\
GPT-5.4 & ChatCut & 2.39 & \textbf{2.70} & 1.91 & 1.87 & \textbf{2.52} & $2.28\pm0.80$ \\
GPT-5.4 & CapCut-Mate & 2.09 & 2.57 & 1.57 & 1.74 & \textbf{2.52} & $2.10\pm0.64$ \\
GPT-5.4 & CutClaw & 1.57 & 1.48 & 1.52 & 1.65 & 1.65 & $1.57\pm0.40$ \\
\bottomrule
\end{tabular}
}
\end{table*}

\subsection{Results}

\begin{figure}[h]
\centering
\includegraphics[width=0.98\linewidth]{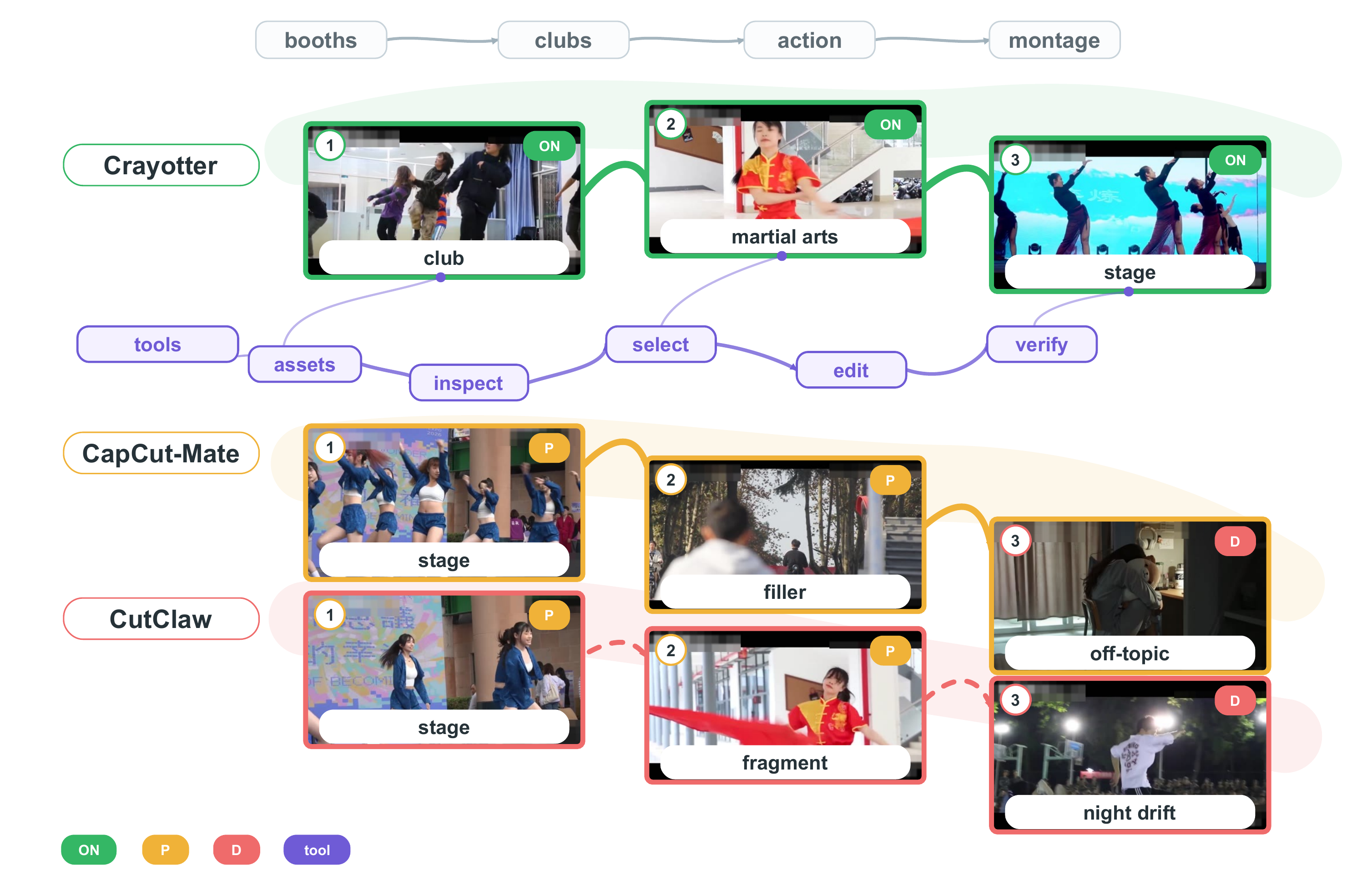}
\caption{Case-level output and tool-trajectory comparison for the campus club fair scenario. The top path gives the target requirements, each row shows the method's output trajectory, and the purple stream marks Crayotter's replayable tool artifacts. ON/P/D denote aligned, partial, and drifted frames.}
\label{fig:case_sample_compare}
\end{figure}

Table~\ref{tab:eval_summary} places Crayotter first overall under both human and GPT-5.4 evaluation. Human raters rank it highest across all dimensions, with ChatCut closest overall. GPT-5.4 ranks ChatCut first in content richness and tied in visual quality, while Crayotter leads the other dimensions and overall. These results compare end-to-end systems rather than controlled backbone substitutions. Figure~\ref{fig:case_sample_compare} illustrates a campus case: Crayotter follows the requested progression and exposes its assets--inspect--select--edit--verify trace, whereas the displayed baselines contain partial or drifted segments.

\section{Discussion}

The evaluation traces complement final-video scores by showing where long-horizon execution succeeds or fails. All 23 Crayotter runs completed under the evaluation protocol, with an average of 221 observable events spanning source analysis, semantic recall, cutting, timeline construction, narration and subtitle alignment, audio processing, and export. These artifacts localize failures that aggregate scores cannot explain, including damaged source intervals, missing requested events, vertical-material leakage, and weak continuity.

Figure~\ref{fig:analysis} connects these diagnostics to the user-facing workbench. Assets, analysis outputs, timeline operations, intermediate renders, and tool events remain inspectable throughout execution, allowing users to trace a failure to the responsible stage and rerun only the affected work. This observe--diagnose--revise loop makes Crayotter a controllable production workspace rather than a prompt-only generator.

\section{Conclusion}

We presented Crayotter, an open-source multimodal multi-agent demo system that makes long-form video editing traceable through coverage-aware material preparation, artifact-grounded research, tool-grounded execution, and reflection. A 23-theme comparison with CapCut-Mate, CutClaw, and ChatCut under human and AI evaluation demonstrates the practical value of this workflow, while resource-aware asynchronous scheduling improves demo usability through concurrent execution, checkpoints, and artifact validation. Crayotter thus turns long-form editing agents from opaque one-shot generators into inspectable, controllable production workspaces.

\section*{Ethics Statement}
\label{app:ethics}

Media shown in the paper and used for evaluation were manually curated from public-domain, openly licensed, or otherwise reuse-authorized sources; private or restricted content was excluded. Public accessibility was not treated as authorization, and users remain responsible for the licensing, attribution, privacy, and permitted use of uploaded or retrieved media.

Three human annotators evaluated generated videos with the same five-dimensional rubric; no demographic or sensitive attributes were collected, and ratings are reported in aggregate or under anonymous identifiers. GPT-5.4 was applied uniformly as an auxiliary judge, with its results reported separately from human judgments rather than treated as ground truth. Because automated retrieval and editing may raise copyright, privacy, or misuse risks, users should verify permissions and review all retrieved assets and generated edits before publication.

\bibliography{sample-bibliography}

\appendix

\section{Extended System Architecture}
\label{app:architecture}

\begin{figure*}[t]
  \centering
  \includegraphics[width=\textwidth]{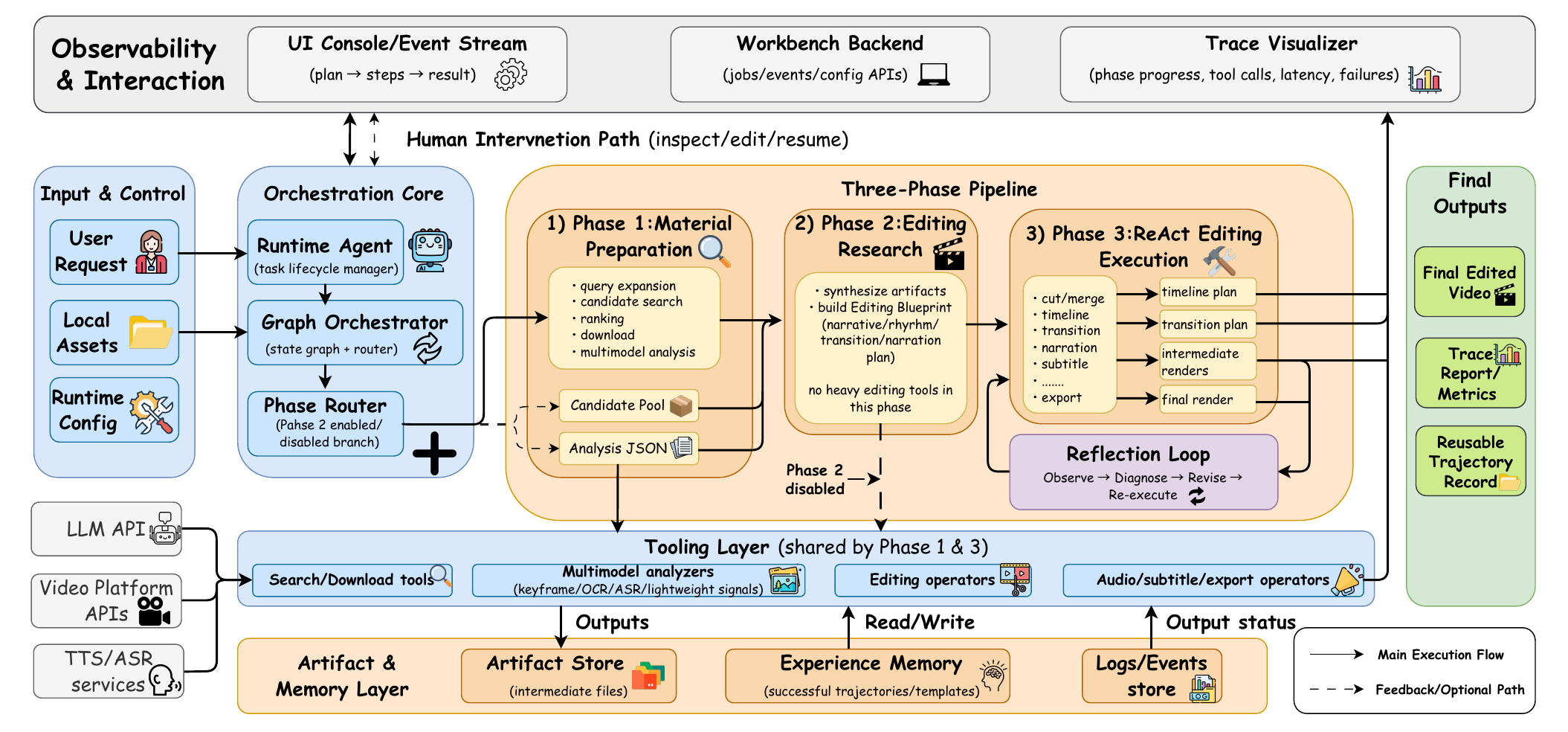}
\caption{Full Crayotter architecture, including the three-phase pipeline, observable artifacts, runtime services, and reusable trajectory logs.}
  \label{fig:overview}
\end{figure*}

Figure~\ref{fig:overview} details the full system architecture and artifact flow across the three editing phases and observable runtime.

\section{Case-Level Evaluation Scores}
\label{app:case_scores}

Tables~\ref{tab:app_human_case_scores} and~\ref{tab:app_ai_case_scores} report case-level scores for all 23 themes and all four systems.
Human overall scores average the five dimensions for each annotator--case--method tuple; GPT-5.4 scores report the five dimensions and their mean.

\begin{table*}[!h]
\centering
\scriptsize
\setlength{\tabcolsep}{2.0pt}
\caption{Case-level human overall scores. H1--H3 denote the annotators; Mean averages their scores for each case--method pair.}
\label{tab:app_human_case_scores}
\resizebox{\textwidth}{!}{%
\begin{tabular}{ll*{16}{c}}
\toprule
Case & Theme & \multicolumn{4}{c}{Crayotter} & \multicolumn{4}{c}{CapCut-Mate} & \multicolumn{4}{c}{CutClaw} & \multicolumn{4}{c}{ChatCut} \\
\cmidrule(lr){3-6}\cmidrule(lr){7-10}\cmidrule(lr){11-14}\cmidrule(lr){15-18}
 & & H1 & H2 & H3 & Mean & H1 & H2 & H3 & Mean & H1 & H2 & H3 & Mean & H1 & H2 & H3 & Mean \\
\midrule
001 & Funny pet fails & 2.4 & 3.2 & 2.8 & 2.80 & 1.6 & 2.0 & 1.8 & 1.80 & 1.4 & 1.0 & 1.6 & 1.33 & 1.8 & 2.6 & 1.8 & 2.07 \\
002 & Lazy cat afternoon & 3.8 & 3.8 & 4.0 & 3.87 & 2.0 & 1.6 & 1.8 & 1.80 & 3.0 & 1.0 & 3.4 & 2.47 & 2.2 & 3.8 & 2.6 & 2.87 \\
003 & Human-pet bonding & 3.0 & 3.4 & 3.6 & 3.33 & 1.2 & 1.0 & 1.2 & 1.13 & 2.0 & 1.0 & 2.2 & 1.73 & 1.0 & 1.0 & 1.4 & 1.13 \\
004 & Library white noise & 4.0 & 3.8 & 4.2 & 4.00 & 1.8 & 1.0 & 2.0 & 1.60 & 2.6 & 1.0 & 2.6 & 2.07 & 3.2 & 4.6 & 3.4 & 3.73 \\
005 & Campus sports & 3.4 & 3.0 & 3.4 & 3.27 & 3.2 & 1.6 & 3.2 & 2.67 & 2.4 & 1.0 & 2.8 & 2.07 & 3.2 & 2.4 & 3.6 & 3.07 \\
006 & Graduation farewell & 3.6 & 4.2 & 3.6 & 3.80 & 3.8 & 2.4 & 4.0 & 3.40 & 2.0 & 1.0 & 2.2 & 1.73 & 3.2 & 3.0 & 3.4 & 3.20 \\
007 & Club fair & 4.8 & 3.6 & 4.8 & 4.40 & 2.6 & 1.0 & 2.8 & 2.13 & 3.0 & 1.0 & 3.2 & 2.40 & 2.4 & 2.8 & 2.4 & 2.53 \\
008 & Campus canteen & 3.8 & 4.0 & 3.6 & 3.80 & 2.2 & 1.0 & 2.6 & 1.93 & 1.8 & 1.0 & 1.6 & 1.47 & 2.6 & 2.6 & 2.4 & 2.53 \\
009 & Road trip & 3.2 & 2.8 & 3.6 & 3.20 & 2.8 & 2.2 & 2.8 & 2.60 & 1.2 & 1.0 & 1.2 & 1.13 & 3.2 & 2.6 & 3.4 & 3.07 \\
010 & City walk & 4.6 & 4.0 & 4.0 & 4.20 & 4.0 & 1.0 & 4.0 & 3.00 & 1.2 & 1.0 & 1.2 & 1.13 & 3.8 & 3.6 & 3.4 & 3.60 \\
011 & Island holiday & 4.2 & 2.8 & 4.2 & 3.73 & 3.6 & 2.4 & 3.4 & 3.13 & 1.8 & 1.0 & 1.8 & 1.53 & 3.4 & 3.4 & 3.6 & 3.47 \\
012 & Mountain trek & 2.6 & 3.0 & 3.0 & 2.87 & 4.4 & 1.0 & 4.6 & 3.33 & 3.8 & 1.0 & 4.0 & 2.93 & 3.8 & 3.2 & 4.0 & 3.67 \\
013 & Ancient-town visit & 3.0 & 3.6 & 3.6 & 3.40 & 3.6 & 2.4 & 3.6 & 3.20 & 1.4 & 1.0 & 1.4 & 1.27 & 5.0 & 4.0 & 4.8 & 4.60 \\
014 & Yili grassland & 2.2 & 3.6 & 2.6 & 2.80 & 2.6 & 2.0 & 2.8 & 2.47 & 3.0 & 1.0 & 3.0 & 2.33 & 4.4 & 3.8 & 4.2 & 4.13 \\
015 & Jiuzhaigou waters & 2.2 & 1.0 & 3.0 & 2.07 & 2.8 & 3.0 & 2.8 & 2.87 & 3.2 & 1.0 & 3.4 & 2.53 & 3.6 & 4.0 & 3.4 & 3.67 \\
016 & Huangshan clouds & 3.8 & 4.4 & 3.8 & 4.00 & 4.0 & 3.2 & 3.8 & 3.67 & 1.6 & 1.0 & 1.4 & 1.33 & 4.4 & 4.4 & 4.0 & 4.27 \\
017 & Guilin landscape & 4.4 & 4.2 & 4.4 & 4.33 & 3.2 & 3.2 & 3.0 & 3.13 & 3.8 & 1.0 & 4.0 & 2.93 & 3.6 & 4.2 & 3.6 & 3.80 \\
018 & Meili sunrise & 3.2 & 3.4 & 3.8 & 3.47 & 3.2 & 2.6 & 3.0 & 2.93 & 1.0 & 1.0 & 1.2 & 1.07 & 3.6 & 3.0 & 3.2 & 3.27 \\
019 & Sichuan hot pot & 3.2 & 3.0 & 3.2 & 3.13 & 2.4 & 1.4 & 2.6 & 2.13 & 1.0 & 1.0 & 1.0 & 1.00 & 4.0 & 4.0 & 3.8 & 3.93 \\
020 & Cantonese morning tea & 3.0 & 3.2 & 3.8 & 3.33 & 1.8 & 2.8 & 1.6 & 2.07 & 1.0 & 1.0 & 1.2 & 1.07 & 3.2 & 3.0 & 3.4 & 3.20 \\
021 & Beijing roast duck & 2.2 & 3.0 & 2.6 & 2.60 & 2.0 & 3.0 & 2.2 & 2.40 & 1.0 & 1.0 & 1.0 & 1.00 & 3.0 & 2.6 & 3.0 & 2.87 \\
022 & Xinjiang barbecue & 2.6 & 2.8 & 3.2 & 2.87 & 1.0 & 2.0 & 1.0 & 1.33 & 1.6 & 1.0 & 1.6 & 1.40 & 3.0 & 2.2 & 3.4 & 2.87 \\
023 & Jiangnan pastries & 2.6 & 3.4 & 2.8 & 2.93 & 1.0 & 1.8 & 1.2 & 1.33 & 1.0 & 1.0 & 1.2 & 1.07 & 2.6 & 3.2 & 2.6 & 2.80 \\
\midrule
Mean & All cases & 3.30 & 3.36 & 3.55 & 3.40 & 2.64 & 1.98 & 2.69 & 2.44 & 1.99 & 1.00 & 2.10 & 1.70 & 3.23 & 3.22 & 3.25 & 3.23 \\
\bottomrule
\end{tabular}%
}
\end{table*}

\begin{table*}[!h]
\centering
\tiny
\setlength{\tabcolsep}{1.0pt}
\caption{Case-level GPT-5.4 scores. TA, CR, NC, ES, VQ, and O denote theme alignment, content richness, narrative coherence, editing smoothness, visual quality, and their mean, respectively.}
\label{tab:app_ai_case_scores}
\resizebox{\textwidth}{!}{%
\begin{tabular}{ll*{24}{c}}
\toprule
Case & Theme & \multicolumn{6}{c}{Crayotter} & \multicolumn{6}{c}{CapCut-Mate} & \multicolumn{6}{c}{CutClaw} & \multicolumn{6}{c}{ChatCut} \\
\cmidrule(lr){3-8}\cmidrule(lr){9-14}\cmidrule(lr){15-20}\cmidrule(lr){21-26}
 & & TA & CR & NC & ES & VQ & \textbf{O} & TA & CR & NC & ES & VQ & \textbf{O} & TA & CR & NC & ES & VQ & \textbf{O} & TA & CR & NC & ES & VQ & \textbf{O} \\
\midrule
001 & Funny pet fails & 2.0 & 3.0 & 2.0 & 2.0 & 1.0 & \textbf{2.0} & 1.0 & 2.0 & 1.0 & 2.0 & 2.0 & \textbf{1.6} & 2.0 & 1.0 & 1.0 & 2.0 & 1.0 & \textbf{1.4} & 2.0 & 4.0 & 2.0 & 2.0 & 2.0 & \textbf{2.4} \\
002 & Lazy cat afternoon & 3.0 & 3.0 & 2.0 & 2.0 & 3.0 & \textbf{2.6} & 2.0 & 3.0 & 1.0 & 1.0 & 2.0 & \textbf{1.8} & 1.0 & 2.0 & 1.0 & 1.0 & 1.0 & \textbf{1.2} & 2.0 & 2.0 & 2.0 & 2.0 & 3.0 & \textbf{2.2} \\
003 & Human-pet bonding & 2.0 & 2.0 & 1.0 & 2.0 & 2.0 & \textbf{1.8} & 1.0 & 1.0 & 1.0 & 1.0 & 2.0 & \textbf{1.2} & 1.0 & 1.0 & 1.0 & 2.0 & 2.0 & \textbf{1.4} & 1.0 & 1.0 & 1.0 & 1.0 & 2.0 & \textbf{1.2} \\
004 & Library white noise & 3.0 & 2.0 & 2.0 & 2.0 & 2.0 & \textbf{2.2} & 1.0 & 2.0 & 1.0 & 1.0 & 2.0 & \textbf{1.4} & 2.0 & 1.0 & 2.0 & 2.0 & 2.0 & \textbf{1.8} & 2.0 & 3.0 & 2.0 & 2.0 & 2.0 & \textbf{2.2} \\
005 & Campus sports & 2.0 & 2.0 & 2.0 & 2.0 & 1.0 & \textbf{1.8} & 2.0 & 2.0 & 1.0 & 1.0 & 2.0 & \textbf{1.6} & 2.0 & 1.0 & 2.0 & 2.0 & 1.0 & \textbf{1.6} & 2.0 & 2.0 & 2.0 & 2.0 & 2.0 & \textbf{2.0} \\
006 & Graduation farewell & 3.0 & 2.0 & 2.0 & 2.0 & 2.0 & \textbf{2.2} & 3.0 & 3.0 & 2.0 & 2.0 & 2.0 & \textbf{2.4} & 2.0 & 1.0 & 2.0 & 2.0 & 1.0 & \textbf{1.6} & 2.0 & 3.0 & 2.0 & 2.0 & 2.0 & \textbf{2.2} \\
007 & Club fair & 4.0 & 5.0 & 3.0 & 3.0 & 4.0 & \textbf{3.8} & 2.0 & 2.0 & 1.0 & 1.0 & 2.0 & \textbf{1.6} & 2.0 & 3.0 & 2.0 & 2.0 & 3.0 & \textbf{2.4} & 1.0 & 1.0 & 1.0 & 1.0 & 2.0 & \textbf{1.2} \\
008 & Campus canteen & 2.0 & 2.0 & 2.0 & 2.0 & 1.0 & \textbf{1.8} & 2.0 & 2.0 & 1.0 & 1.0 & 2.0 & \textbf{1.6} & 2.0 & 1.0 & 2.0 & 2.0 & 1.0 & \textbf{1.6} & 2.0 & 2.0 & 1.0 & 1.0 & 2.0 & \textbf{1.6} \\
009 & Road trip & 3.0 & 3.0 & 2.0 & 2.0 & 3.0 & \textbf{2.6} & 2.0 & 3.0 & 2.0 & 2.0 & 3.0 & \textbf{2.4} & 2.0 & 2.0 & 1.0 & 2.0 & 2.0 & \textbf{1.8} & 2.0 & 2.0 & 1.0 & 1.0 & 2.0 & \textbf{1.6} \\
010 & City walk & 4.0 & 3.0 & 2.0 & 2.0 & 2.0 & \textbf{2.6} & 3.0 & 4.0 & 2.0 & 3.0 & 4.0 & \textbf{3.2} & 1.0 & 1.0 & 1.0 & 1.0 & 2.0 & \textbf{1.2} & 3.0 & 4.0 & 2.0 & 2.0 & 4.0 & \textbf{3.0} \\
011 & Island holiday & 3.0 & 2.0 & 2.0 & 3.0 & 3.0 & \textbf{2.6} & 2.0 & 3.0 & 2.0 & 2.0 & 3.0 & \textbf{2.4} & 2.0 & 1.0 & 2.0 & 2.0 & 2.0 & \textbf{1.8} & 2.0 & 3.0 & 2.0 & 1.0 & 3.0 & \textbf{2.2} \\
012 & Mountain trek & 2.0 & 2.0 & 2.0 & 2.0 & 2.0 & \textbf{2.0} & 1.0 & 2.0 & 2.0 & 2.0 & 2.0 & \textbf{1.8} & 1.0 & 1.0 & 1.0 & 1.0 & 2.0 & \textbf{1.2} & 2.0 & 2.0 & 2.0 & 3.0 & 3.0 & \textbf{2.4} \\
013 & Ancient-town visit & 3.0 & 3.0 & 2.0 & 2.0 & 2.0 & \textbf{2.4} & 4.0 & 4.0 & 3.0 & 3.0 & 3.0 & \textbf{3.4} & 2.0 & 2.0 & 2.0 & 2.0 & 2.0 & \textbf{2.0} & 3.0 & 3.0 & 2.0 & 2.0 & 2.0 & \textbf{2.4} \\
014 & Yili grassland & 4.0 & 3.0 & 3.0 & 4.0 & 4.0 & \textbf{3.6} & 3.0 & 3.0 & 2.0 & 2.0 & 3.0 & \textbf{2.6} & 2.0 & 3.0 & 2.0 & 2.0 & 2.0 & \textbf{2.2} & 4.0 & 4.0 & 3.0 & 3.0 & 4.0 & \textbf{3.6} \\
015 & Jiuzhaigou waters & 2.0 & 2.0 & 2.0 & 3.0 & 2.0 & \textbf{2.2} & 2.0 & 2.0 & 1.0 & 2.0 & 3.0 & \textbf{2.0} & 2.0 & 2.0 & 1.0 & 1.0 & 1.0 & \textbf{1.4} & 4.0 & 4.0 & 3.0 & 3.0 & 4.0 & \textbf{3.6} \\
016 & Huangshan clouds & 4.0 & 4.0 & 3.0 & 2.0 & 3.0 & \textbf{3.2} & 4.0 & 4.0 & 3.0 & 3.0 & 3.0 & \textbf{3.4} & 1.0 & 1.0 & 1.0 & 1.0 & 1.0 & \textbf{1.0} & 4.0 & 4.0 & 3.0 & 3.0 & 4.0 & \textbf{3.6} \\
017 & Guilin landscape & 4.0 & 3.0 & 3.0 & 3.0 & 4.0 & \textbf{3.4} & 3.0 & 3.0 & 2.0 & 2.0 & 3.0 & \textbf{2.6} & 2.0 & 3.0 & 2.0 & 2.0 & 3.0 & \textbf{2.4} & 4.0 & 3.0 & 3.0 & 3.0 & 3.0 & \textbf{3.2} \\
018 & Meili sunrise & 2.0 & 2.0 & 2.0 & 2.0 & 2.0 & \textbf{2.0} & 3.0 & 2.0 & 2.0 & 2.0 & 3.0 & \textbf{2.4} & 1.0 & 1.0 & 1.0 & 2.0 & 2.0 & \textbf{1.4} & 3.0 & 3.0 & 2.0 & 2.0 & 3.0 & \textbf{2.6} \\
019 & Sichuan hot pot & 3.0 & 3.0 & 2.0 & 2.0 & 3.0 & \textbf{2.6} & 2.0 & 3.0 & 1.0 & 2.0 & 3.0 & \textbf{2.2} & 1.0 & 1.0 & 1.0 & 1.0 & 2.0 & \textbf{1.2} & 4.0 & 4.0 & 3.0 & 2.0 & 3.0 & \textbf{3.2} \\
020 & Cantonese morning tea & 2.0 & 2.0 & 2.0 & 1.0 & 1.0 & \textbf{1.6} & 1.0 & 2.0 & 1.0 & 1.0 & 2.0 & \textbf{1.4} & 1.0 & 1.0 & 2.0 & 1.0 & 1.0 & \textbf{1.2} & 1.0 & 2.0 & 1.0 & 1.0 & 1.0 & \textbf{1.2} \\
021 & Beijing roast duck & 2.0 & 2.0 & 1.0 & 1.0 & 1.0 & \textbf{1.4} & 1.0 & 2.0 & 1.0 & 1.0 & 2.0 & \textbf{1.4} & 1.0 & 1.0 & 1.0 & 1.0 & 1.0 & \textbf{1.0} & 1.0 & 1.0 & 1.0 & 1.0 & 1.0 & \textbf{1.0} \\
022 & Xinjiang barbecue & 2.0 & 2.0 & 2.0 & 2.0 & 2.0 & \textbf{2.0} & 2.0 & 3.0 & 2.0 & 2.0 & 3.0 & \textbf{2.4} & 2.0 & 2.0 & 2.0 & 2.0 & 1.0 & \textbf{1.8} & 2.0 & 2.0 & 1.0 & 1.0 & 2.0 & \textbf{1.6} \\
023 & Jiangnan pastries & 3.0 & 3.0 & 3.0 & 2.0 & 2.0 & \textbf{2.6} & 1.0 & 2.0 & 1.0 & 1.0 & 2.0 & \textbf{1.4} & 1.0 & 1.0 & 2.0 & 2.0 & 2.0 & \textbf{1.6} & 2.0 & 3.0 & 2.0 & 2.0 & 2.0 & \textbf{2.2} \\
\midrule
Mean & All cases & 2.78 & 2.61 & 2.13 & 2.17 & 2.26 & \textbf{2.39} & 2.09 & 2.57 & 1.57 & 1.74 & 2.52 & \textbf{2.10} & 1.57 & 1.48 & 1.52 & 1.65 & 1.65 & \textbf{1.57} & 2.39 & 2.70 & 1.91 & 1.87 & 2.52 & \textbf{2.28} \\
\bottomrule
\end{tabular}%
}
\end{table*}

\section{Inter-Annotator Reliability}
\label{app:agreement}

\begingroup
\begin{center}
\small
\captionof{table}{Inter-annotator reliability for five dimensions across 92 case--method outputs. ICC(A,3) is a two-way random-effects, absolute-agreement coefficient for the mean of three annotators; 95\% CIs are percentile intervals from 20,000 cluster-bootstrap samples over themes.}
\label{tab:app_icc}

\resizebox{\linewidth}{!}{%
\begin{tabular}{lcc}
\toprule
\textbf{Dimension} & \textbf{ICC(A,3)} & \textbf{95\% CI} \\
\midrule
Theme alignment & 0.773 & 0.678--0.843 \\
Content richness & 0.794 & 0.684--0.867 \\
Narrative coherence & 0.810 & 0.722--0.870 \\
Editing smoothness & 0.775 & 0.685--0.835 \\
Visual quality & 0.834 & 0.750--0.889 \\
\midrule
Overall & 0.857 & 0.776--0.911 \\
\bottomrule
\end{tabular}%
}
\end{center}
\endgroup

\clearpage

\end{document}